\documentclass[conference]{IEEEtran}
\usepackage{times}

\usepackage[numbers,]{natbib}
\usepackage{multicol}
\usepackage{graphicx}
\usepackage{subfigure}
\usepackage[bookmarks=true]{hyperref}
\usepackage{svg}

\usepackage{amsmath}
\usepackage{amssymb}
\usepackage{textcomp}
\usepackage{xcolor}
\usepackage{pgfplots}
\usepgfplotslibrary{groupplots}
\usepackage{tikz}
\usepackage{tikzscale}
\usepackage{cleveref}
\usepackage{subcaption}
\usepackage{float}
\usepackage{multirow}
\usepackage{makecell}
\usepackage{caption}
\usepackage{mathtools}
\usepackage{algorithm}
\usepackage[indLines=true,noEnd=true,rightComments=false]{algpseudocodex} 
\usepackage{siunitx}
\usepackage{comment}
\usepackage{booktabs}
\usepackage{alphalph}
\usepackage{wrapfig}
\usepackage{bbm}

\crefname{section}{Sec.}{Section}
\Crefname{section}{Sec.}{Section}
\crefname{figure}{Fig.}{Figure}
\Crefname{figure}{Fig.}{Figure}
\crefname{table}{Table.}{Table}
\Crefname{table}{Table.}{Table}
\crefname{equation}{}{}
\Crefname{equation}{Equation}{Equations}

\newcommand{\ie}{\textit{i}.\textit{e}., }
\newcommand{\eg}{\textit{e}.\textit{g}., }

\newcommand{\nspack}{NeSyPack}
\newcommand{\myparagraph}[1]{\vspace{3pt} \noindent \textbf{#1} \ }

\begin{document}

\title{\LARGE \bf
NeSyPack:\\A Neuro-Symbolic Framework for Bimanual Logistics Packing}

\author{
  Bowei Li$^{*, 1}$, Peiqi Yu$^{*,1}$, Zhenran Tang$^{1}$, Han Zhou$^{1}$, Yifan Sun$^{1}$, Ruixuan Liu$^{1}$ and Changliu Liu$^{1}$\\
  $^*$Equal Contribution, $^1$Carnegie Mellon University \\
}



%

\maketitle

\begin{abstract}
This paper presents \nspack{}, a neuro-symbolic framework for bimanual logistics packing.
Our \nspack{} combines data-driven models and symbolic reasoning to build an explainable hierarchical framework that is generalizable, data-efficient, and reliable.
It decomposes a task into subtasks via hierarchical reasoning, and further into atomic skills managed by a symbolic skill graph. The graph selects skill parameters, robot configurations, and task-specific control strategies for execution.
This modular design enables robustness, adaptability, and efficient reuse—outperforming end-to-end models that require large-scale retraining.
Using \nspack{}, our team won the \textit{First Prize} in the \textit{What Bimanuals Can Do (WBCD)} competition at the 2025 IEEE International Conference on Robotics \& Automation (ICRA).
\end{abstract}

\IEEEpeerreviewmaketitle

\section{Introduction}

Logistics packing is a crucial task in the warehouse industry, which requires personnel to select the appropriate items and pack them into a shipping box.
Existing works assume objects have smooth surfaces and use suction grippers \cite{amazon_sparrow,siemens,zhang2025physics}.
However, objects with uneven or air-leaking surfaces cannot be manipulated by suction grippers.
Recent work \cite{hudson2025stow} integrates visual and tactile feedback to design an innovative two-finger gripper, enabling a robot to pick items and stow them into shelves.
However, it considers single-arm operation, which remains difficult to perform complex tasks, \eg sealing the packing box.
Recent advancements in humanoid robots bring attention to bimanual systems, offering higher efficiency,  larger workspaces, and more dexterous manipulation capabilities \cite{huang2025apexmr}.
However, it remains under exploration regarding how to enable the robots to quickly learn packing skills with a wide variety of different objects. 
One widely studied approach is vision-language-action models (VLA) \cite{black2024pi_0,kim2024openvla,intelligence2025pi_}, which enable robots to perform a wide variety of different tasks, \eg folding clothes, hanging cups, wiping tables, etc.
\cite{kim2024openvla} uses perception models \cite{oquab2023dinov2,zhai2023sigmoid} to extract visual features and uses large-language models (LLM) \cite{touvron2023llama} to encode extracted features into a hidden state.
They then train an action policy, \ie an action decoder, which generates robot action commands from the hidden state.
\cite{black2024pi_0,intelligence2025pi_} decouples the VLA pipeline into a larger vision-language model (VLM) backbone \cite{beyer2024paligemma} and a smaller action expert model for generating robot commands from the VLM output. 
However, VLA-based methods are unexplainable and require a massive amount of high-quality data to achieve decent performance and generalizability.
And it is difficult, if not impossible, to obtain massive high-quality data for industrial settings. 
\cite{siemens} uses a model-based approach, in which they train a neural network model to extract the geometric features of the objects, and manually design a policy for grasping, \ie using the suction gripper to grasp from the center point of the visible flat surface.
However, due to the specialized gripper and the single unified action policy, it is difficult to generalize to more complex tasks.

This paper presents \nspack{},  a neuro-symbolic framework \cite{10.5555/3326943.3327039} for bimanual logistics packing that enables robots to efficiently learn and perform packing tasks involving a diverse range of objects. 
Our approach combines data-driven models and symbolic reasoning across multiple levels. 
First, perceptive features are represented symbolically---for example, tracking the edges of deformable objects---while the perception model that maps raw visual and depth data to these features is learned. Second, manipulation skills in Cartesian space, such as determining hand trajectories based on perceptual input, are acquired through human demonstrations. These range from simple frame-relative offsets (e.g., for rigid object pick-and-place) to neural-network-based policies (e.g., for deformable object manipulation). Thirdly, while the manipulation skills in the Cartesian space are learned, the supportive robot arm motions are calculated symbolically, accounting for all potential objectives and constraints in tracking the Cartesian space motion.
Finally, all learned models, control policies, and symbolic computation modules are organized into a unified neuro-symbolic structure---the skill graph. An online symbolic planner leverages this framework to select appropriate representations, perception models, and motion policies to fulfill high-level task specifications.
By injecting symbolic reasoning into neural models, \nspack{} is explainable and demonstrates strong generalizability, high data efficiency, and promising reliability.
We deploy \nspack{} in the \textit{What Bimanuals Can Do (WBCD)} competition \footnote{\url{https://2025.ieee-icra.org/competitions/\#wbcd}} at the 2025 IEEE International Conference on Robotics \& Automation (ICRA).
Due to the strong generalizability and data efficiency, we were the only team able to fully deploy our system to a completely new competition setup.
Upon successful transfer, our system showed high reliability and eventually won the \textbf{First Prize}.


\begin{figure*}
    \centering
    \includegraphics[width=\linewidth]{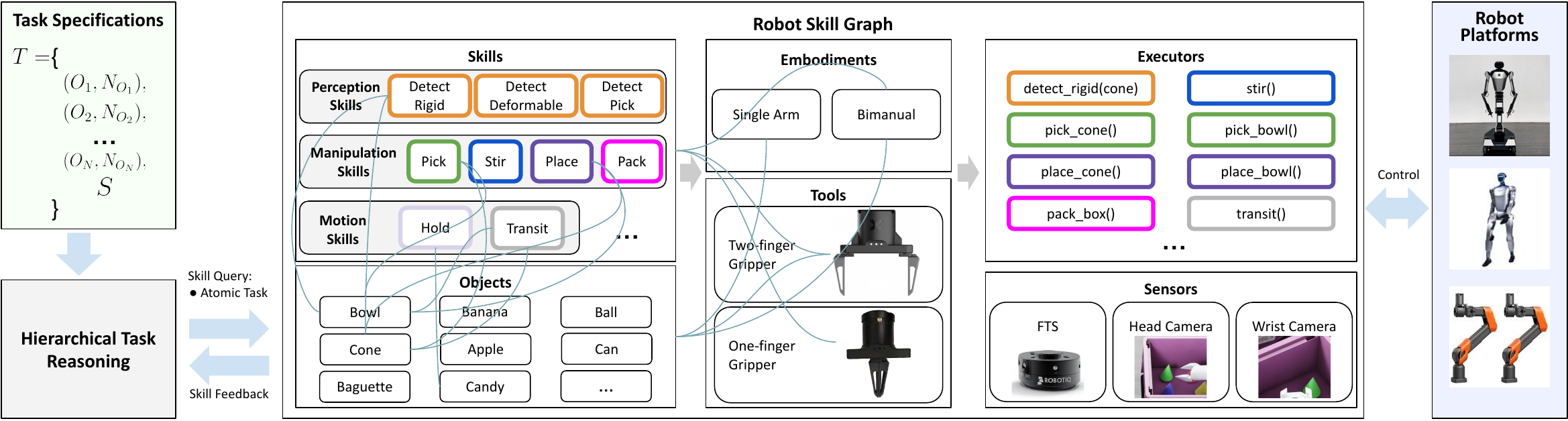}
    \caption{ \textbf{An overview of \nspack{}}. Given the packing task $T$, the HTR decomposes the long-horizon task into smaller atomic tasks and queries the robot skill graph to execute individual atomic tasks using the appropriate skills.
    }
    \label{fig:system_overview}
    \vspace{-15pt}
\end{figure*}

\section{\nspack{}: A Neuro-Symbolic Framework for Bimanual Logistics Packing}

This paper addresses the logistics packing problem, in which the system needs to pick designated items from storage bins, place them into the shipping box, and properly enclose the shipping box.
A full packing task can be denoted as $T=[(O_1, N_{O_1}), (O_2, N_{O_2}), \dots, (O_N, N_{O_N}), S]$, which consists of $N$ different types of items to be packed followed by a box sealing task $S$.
$N_{O_i}$ indicates the quantity needed for item $O_i$.
\Cref{fig:system_overview} illustrates the system overview of \nspack{}, which includes two major components: 1) the hierarchical task reasoning (HTR) module, and 2) the robot skill graph.
Given the packing task, the HTR decomposes the long-horizon task $T$ into subtasks and supervises the overall task progress.
Based on the monitored task status, the skill graph extracts the symbolic representation, and then reasons for the best-suited action policies, and commands the robot for execution.

\subsection{Hierarchical Task Reasoning}\label{sec:htr}
\begin{figure}
    \centering
    \includegraphics[width=\linewidth]{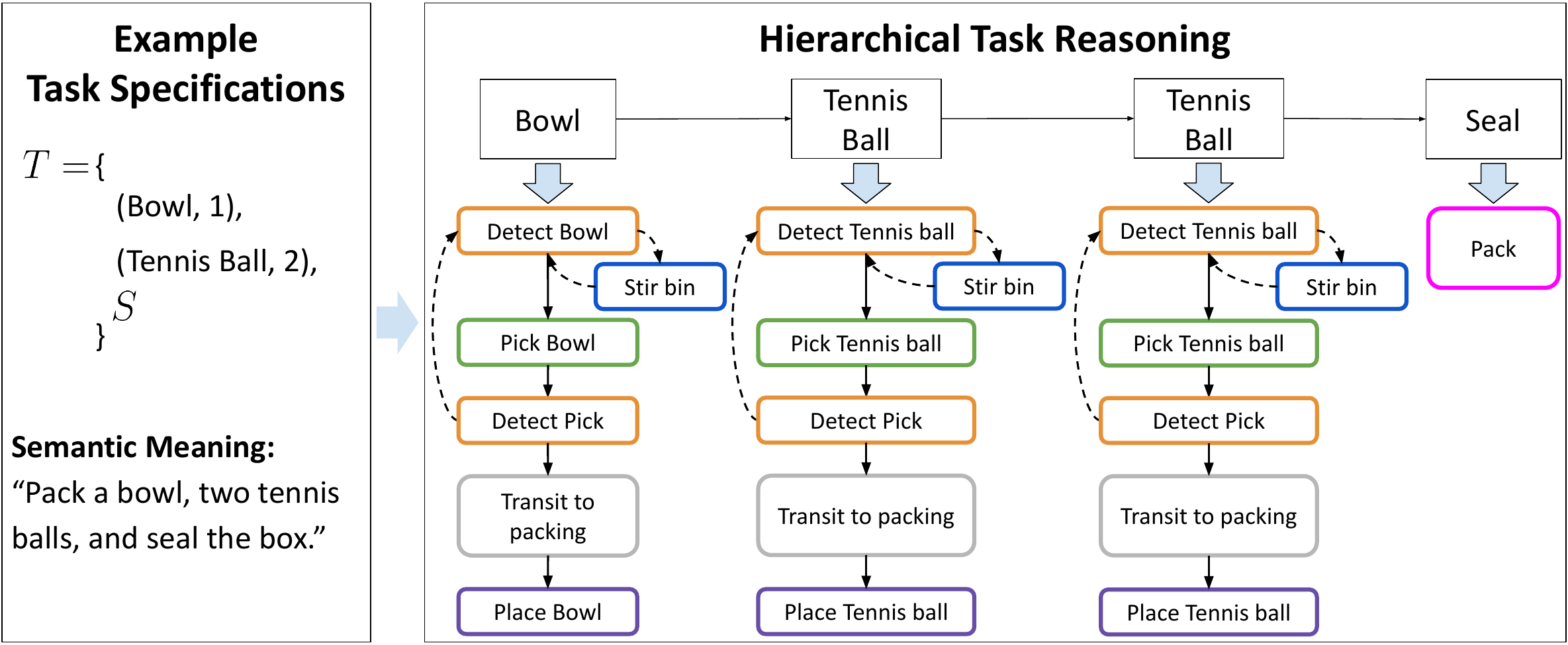}
    \caption{ \textbf{Illustration of the HTR on an example task}. Solid arrows: determined task plans. Dashed arrows: tentative task plans subject to change.
    }
    \label{fig:htr}
    \vspace{-15pt}
\end{figure}

As illustrated in \cref{fig:htr}, given the packing task, the HTR module decomposes the long-horizon task $T$ into subtasks $o_1^1, o_1^2, \dots, o_1^{N_{O_1}}, o_2^1, \dots, o_N^{N_{O_N}}, S$.
For instance, for the task `Pack one bowl, two tennis balls, and seal the box'' as shown in \cref{fig:htr}, the decomposed subtasks are 1) $o_1^1$: pack a bowl, 2) $o_2^1$: pack a tennis ball, 3) $o_2^2$: pack a tennis ball, and 4) $S$: seal the box.
Given each subtask, HTR further decomposes it into smaller atomic tasks.
For instance, for the subtask $o_1^1$ (\ie packing a bowl in \cref{fig:htr}), a decomposed atomic task plan would be 1) detecting the 3D pose of the bowl, 2) picking up the bowl, 3) detecting successful pick, 4) transferring the bowl to the packing box, and 5) placing the bowl into the packing box.
Essentially, HTR decomposes a long-horizon task into a step-by-step execution plan \cite{10552885}, in which each individual atomic task is within the robot's capability.
Due to the reduced horizon, each atomic task can be easily parameterized or learned.
To physically execute the atomic task, HTR queries the skill graph as shown in \cref{fig:system_overview}.

In addition, HTR monitors the overall task progress and adjusts the nominal plan when necessary \cite{chen2024automating}.
As illustrated in \cref{fig:htr}, certain atomic tasks—\eg ``Stir bin''—are not required for every subtask $o_i^j$.
Therefore, they are omitted from the default plan but can be dynamically inserted as failure recovery policies when perception fails, as indicated by the dashed arrows.

\subsection{Skill Graph}\label{sec:skill_graph}


As illustrated in \cref{fig:system_overview}, the skill graph takes an atomic task from HTR as input, then extracts its symbolic representation and reasons for the best-suited action policy. 
Specifically, it identifies the target object and required skill for the atomic task, determines the appropriate embodiment and tool, and outputs grounded execution functions as control commands for the robot.
The skill graph integrates both learned models and symbolic components across six interdependent elements that facilitate structured and adaptive task execution:
\begin{enumerate}
    \item
    \textit{Skills}: Represent high-level, semantic robot operations that facilitate intuitive task planning, including perception skills such as ``Detect Object'', manipulation skills such as ``Pick'', ``Place'', ``Stir'', and ``Pack'', and motion skills such as ``Hold'', and ``Transit'', etc.
    \item 
    \textit{Objects}: Items to be packed, each described by their physical properties—such as size, shape, weight, and rigidity—as well as perceptive features that guide manipulation. For example, spherical objects are represented using position only, while cylindrical objects (e.g., cans) additionally include an upright flag derived from pose estimation. These features directly influence the selection of manipulation strategies: a can with the upright flag is grasped from the top, whereas a tilted can is grasped perpendicularly to its longitudinal axis.

    \item 
    \textit{Embodiments}: Define supported robot configurations, including single-arm and dual-arm.
    \item 
    \textit{Tools}: Specify the available end-of-arm tool (EOAT) to use, \eg ``two-finger gripper'' and ``one-finger gripper".
    \item 
    \textit{Executors}: Link each skill-object set to executable functions, generating execution policy (pre-trained) while incorporating embodiment, tools, and feedback sensors.
    \item 
    \textit{Sensors}: Specify sensors that provide feedback.
\end{enumerate}

The components within the skill graph are highly interdependent. For example, as shown in \cref{fig:system_overview}, given an atomic task ``Pick Bowl'' by HTR, the skill graph extracts the required skill being ``pick'' and the object being ``bowl'', and determines ``bimanual'' embodiment, uses one-finger gripper, integrates the head and wrist cameras for feedback.
Based on the selections, it then executes ``pick\_bowl()'' to generate robot motions and perform the task. 
This structured approach has several advantages: 1) its modular design allows robots to reuse skills, enabling efficient adaptation to diverse tasks without end-to-end re-training on a massive amount of data; 2) it stores the relationships between skills and hardware resources, enabling efficient and robust grounding of robot skills; 3) it supports efficient expansion of skill sets without affecting existing skills.

In skill graph, we define a robot skill as a parameterized, goal-directed capability that transforms the world state or change the robot's knowledge about the world, to progress toward a task-specific objective. With this definition, we divide robot skills into three categories: perception skills, manipulation skills, and motion skills.

\myparagraph{Perception Skills}
Perception skills are modular, task-agnostic components that update the robot’s internal world model and provide symbolic perceptive features. Their design ensures flexible integration into a wide variety of object types and manipulation contexts.

1) Detect Rigid:
For rigid objects with known geometries (e.g., cans and bowls), this skill uses YOLOV11 \cite{jocher2023yolov11} for instance segmentation on RGB-D input, then back-projects masks into 3D to form dense point clouds.
Semantic priors (e.g., removing object points from the box point cloud) help reduce occlusion artifacts.
Each object and box point cloud is registered to its CAD model using ICP with multiple initial guesses.
The resulting 6D poses are transformed into the robot’s base frame for downstream manipulation.

2) Detect Deformable:
For deformable or unmodeled objects (e.g., clothing) without CAD models, the system bypasses model registration and directly analyzes the segmented point cloud.
It detects boundaries and computes prominent edges using both masks and depth data.
By ranking clean, accessible edges, it generates candidate grasp points and orientations to guide manipulation of soft or irregular items.

3) Detect Pick:
This skill monitors if a pick is successful. It determines success by comparing the gripper pose and the perceived object pose from point cloud data. Spatial alignment is used to verify that the object has been securely picked.


Modular perception skills enable our system to adapt strategies based on object type and scene context. This design ensures extensibility for new objects or sensors and provides reliable, context-aware outputs for long-horizon tasks.

\myparagraph{Manipulation Skills}
The manipulation skills represent structured, reusable action primitives that physically interact with objects through coordinated robot motion and gripper control. These include task-level operations such as ``Pick'', ``Place'', ``Stir'', and ``Pack''. Each skill is parameterized and modular, enabling generalization across tasks and platforms without retraining. 
To promote generalization and robustness of skills, we first do motion planning in Cartesian space. 
Specifically, both hard constraints (\eg kinematic reachability, collision avoidance) and soft constraints (\eg preferred grasp orientation, trajectory smoothness) are integrated into motion planning as labels for waypoints\cite{Kuffner2000-hn,cfs}. And then we map these labeled Cartesian waypoints to configuration space.
This hybrid constraint formulation enables robust and interpretable control even in cluttered or dynamic environments. 

Specifically, we customize each individual manipulation skill as follows. A ``Pick'' skill typically consists of moving to a pre-grasp pose, approaching and executing the grasp, and lifting the object. 
Specifically, we customize the ``Pick'' skill by object type through parameter templates:
\begin{itemize}
    \item \textbf{Small rigid objects:} Center-based grasp poses are selected and evaluated across multiple orientations to avoid collisions.
    \item \textbf{Cylindrical objects:} Grasp alignment adapts to object pose—horizontal objects are grasped perpendicular to their major axis, upright ones use a center-based strategy.
    \item \textbf{Large or heavy objects:} Diagonal corner grasps using two arms improve load distribution and enhance stability.
    \item \textbf{Deformable objects:} Grasp candidates are generated by analyzing object boundaries and edges from the segmented point cloud, without relying on CAD models.

\end{itemize}
A ``Place'' skill involves choosing a target location, moving to that location, and opening the gripper to place the object.
Other manipulation tasks such as ``Stir'' and ``Pack'' may follow similar structured sequences, but are often executed through teleoperation or learned via imitation learning as a bin-pose-conditioned neural policy due to their complexity or variability.

\myparagraph{Motion Skills}
Unlike manipulation skills, the motion skills do not physically interact with the environment.
The skill graph includes two fundamental motion skills: ``Hold'' and ``Transit''. 
The ``Transit'' skill is responsible for collision-free movements between key poses. 
In our system, we use RRT-Connect \cite{Kuffner2000-hn} to generate collision-free dual-arm motions.
The ``Hold'' skill pauses the robot's motion for a specified time.
It is used to keep the robot static during critical decision-making, sensing, or coordination phases, ensuring accurate downstream actions and overall system stability.



\section{Results}

\begin{figure}
    \centering
    \includegraphics[width=\linewidth]{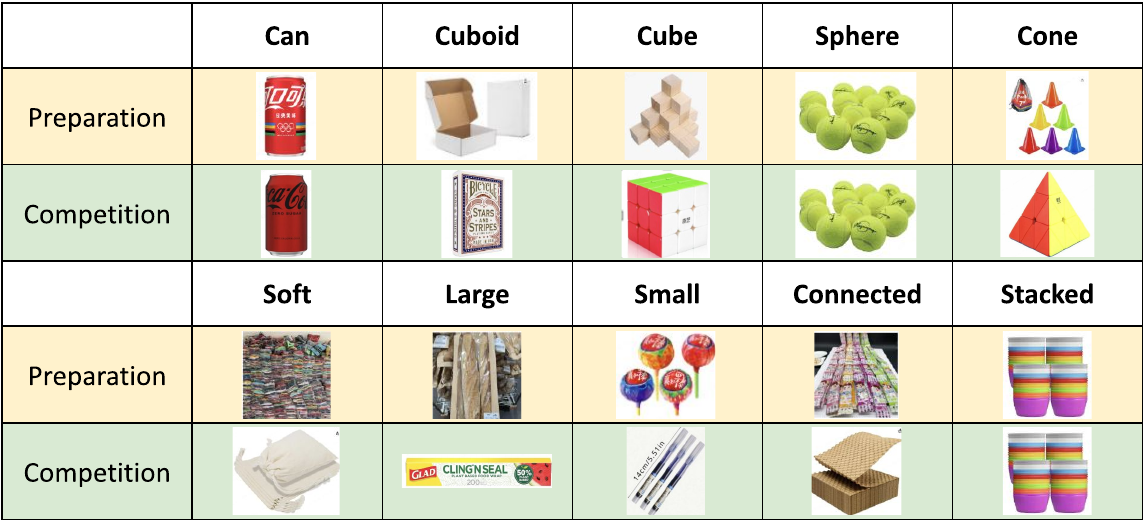}
    \caption{ \textbf{Packing Object List}. Yellow: Objects available during preparation. Green: Items used during the WBCD competition.
    }
    \label{fig:object-list}
    \vspace{-15pt}
\end{figure}

We consider a diverse range of objects across ten categories, each posing different manipulation challenges as shown in \cref{fig:object-list}. 
These include cylinders (\eg tin cans), cubes (\eg wooden blocks), spheres (\eg tennis balls), and soft objects (\eg T-shirts), as well as more complex items such as cuboids and cones that require bimanual handling; tightly stacked or connected items that must be separated using two arms; large flat objects, \eg baguettes, that need to be carefully packed; and small items, \eg candy lollipops, that require precise manipulation. 
To ensure the system is generalizable, only the items shown in the yellow cells were available during development.
The items in the green cells were unveiled three days before the competition.
In addition, the robot hardware was not provided before the competition, and thus, we developed \nspack{} on our own platform and transferred it to the official platform right before the competition.

\subsection{Development on Unitree G1}

\begin{figure}
    \centering
    \subfigure[ Unitree G1]{\includegraphics[width=0.49\linewidth]{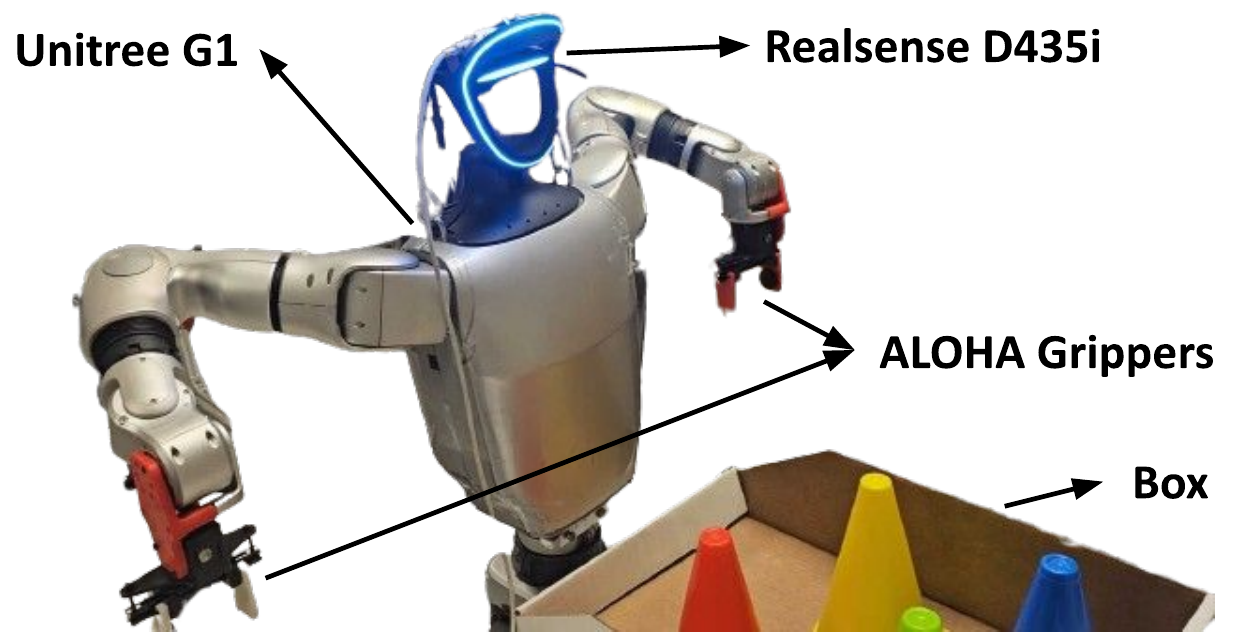}\label{fig:g1}}\hfill
    \subfigure[ Dual Galaxea A1X]{\includegraphics[width=0.49\linewidth]{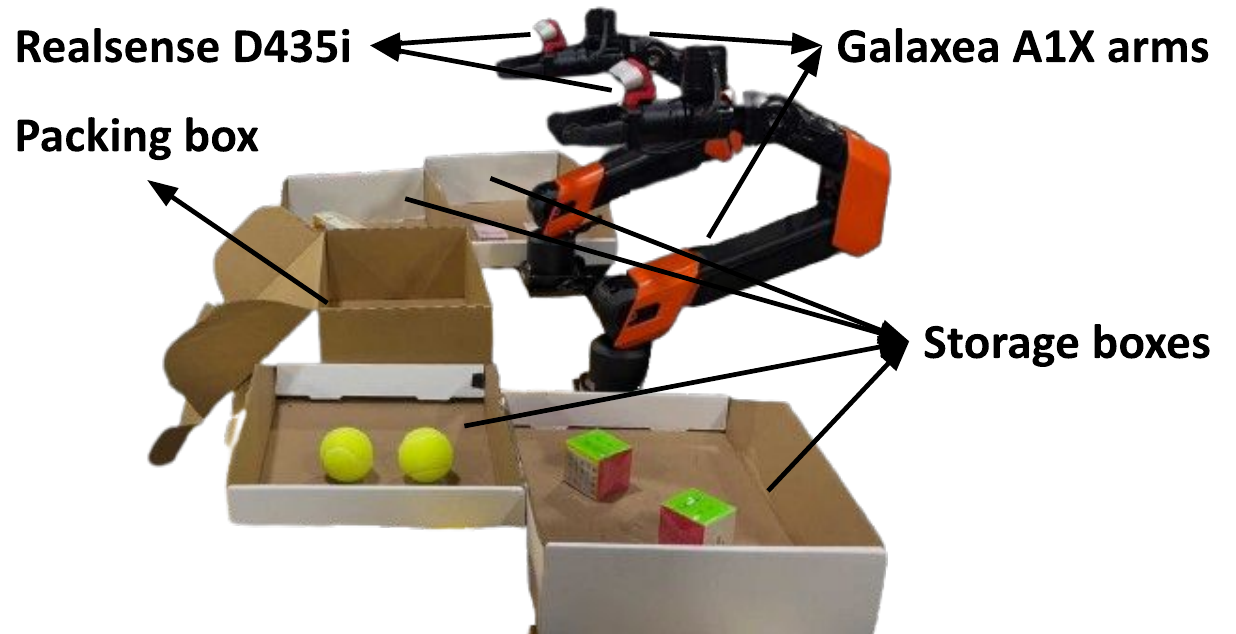}\label{fig:a1x}}\hfill
    \vspace{-5pt}
    \caption{ \textbf{Robot Hardware Setup.} Left: Preparation setup. Right: Deployment (WBCD competition) setup. 
    }
    \label{fig:experiment_setup}
    \vspace{-15pt}
\end{figure}

The proposed \nspack{} was developed on our G1 humanoid as shown in \cref{fig:g1}.
The robot is equipped with an ALOHA gripper on each arm for bimanual manipulation. 
A RealSense D435i depth camera is mounted on the robot's head to provide real-time visual feedback for perception skills. 
The objects to be packed are placed in an open-top storage box located in front of the robot. 
Due to the limited reachability of the G1 arms, only one box is presented in front of the robot, and we manually switch between storage and packing boxes.

We evaluated the system using objects available during preparation as shown in \cref{fig:object-list}. 
Each object was tested for 10 trials, and as shown in \cref{tab:object_successes}, the system achieved high success rates across all objects, with all categories reaching 9/10 or 10/10. 
This indicates that the proposed \nspack{} is reliable for bimanual logistics packing with diverse objects. 

\begin{table}
\centering
\caption{ Packing Success Rates on G1}
\label{tab:object_successes}
\begin{tabular}{lccc}
\toprule
\textbf{Object} & \textbf{Success Count} & \textbf{Total Attempts} & \textbf{Success Rate} \\
\midrule
Cube     & 9 & 10 & 90\% \\
Sphere     & 9 & 10 & 90\% \\
Stacked     & 10 & 10 & 100\% \\
Can      & 9 & 10 & 90\% \\
Cone     & 9 & 10 & 90\% \\
Small & 10 & 10 & 100\% \\
\bottomrule
\end{tabular}
\end{table}

\subsection{WBCD Competition at ICRA 2025}

We participated in the What Bimanuals Can Do (WBCD) competition at ICRA 2025.
\Cref{fig:a1x} illustrates the hardware setup during the competition, which significantly differs from our preparation setup.
The competition setup includes two Galaxea A1X arms, with a RealSense D435i on each of their wrist.
Due to the larger reachability, five boxes were presented to the bimanual system, including four storage boxes and one packing box.
The robot must pick the correct number of items from each storage bin and transport them to the central packing box. 
Our system was evaluated in four official trials. 
The packing success rates across different object types are summarized in \cref{tab:wbcd-success}. 
Videos of our trials are available at \url{http://icontrol.ri.cmu.edu/news/icra2025-wbcd.html}.

\begin{table}
\centering
\caption{ Packing Success Rates in WBCD Competition}
\label{tab:wbcd-success}
\begin{tabular}{lccc}
\toprule
\textbf{Object} & \textbf{Success Count} & \textbf{Total Attempts} & \textbf{Success Rate} \\
\midrule
Cube        & 2  & 2 & 100\%  \\
Sphere      & 5  & 6 & 83.3\% \\
Large   & 2  & 2 & 100\% \\
Cuboid       & 4  & 4 & 100\% \\
Stacked       & 5  & 6 & 83.3\% \\
\bottomrule
\end{tabular}
\vspace{-15pt}
\end{table}

\myparagraph{Generalizability}
Our system supports rapid adaptation to new robot platforms with minimal engineering overhead.
For different robots, all computations involving robot-specific parameters are performed symbolically.
The exact placement of the camera is not critical, since the perceptual features are ultimately transformed into the global coordinate frame for reasoning.
Similarly, the kinematic parameters of the gripper and arm are only used in configuration space planning, which is also handled symbolically. As a result, the system can be easily migrated to new platforms without retraining models or rewriting control logic.

In addition, our method exhibits strong task generalization. The extracted perceptual features and manipulation policies are primarily based on the geometric properties of objects. This allows the system to generalize well to objects with similar shapes—such as comparable sizes, edge structures, or contours—particularly in terms of avoiding collisions and achieving stable grasps.
However, due to the lack of force feedback, the system has limited generalization to variations in surface materials. For example, slippery objects are more likely to be dropped. To address this issue, we added rubber pads with increased friction to the gripper fingertips, which significantly improves grasp robustness and adaptability.

\myparagraph{Data Efficiency}
Perception models were adapted in under one hour on an RTX 4060 laptop using only a handful of labeled images per object category; manipulation skills are acquired via one-shot key-frame demonstrations specifying key poses and gripper actions.

\myparagraph{Reliability}
Our original G1-based pipeline achieved over 90\% task success in controlled laboratory evaluations. 
When deployed on the dual Galaxea A1X platform, this level of performance was preserved as shown in \cref{tab:wbcd-success}, culminating in a \textbf{first-place} finish in the WBCD competition.

\myparagraph{Extensibility}
Our neuro-symbolic framework allows seamless integration of new perception and manipulation skills for unseen object types. These can be added as standalone modules or by augmenting existing ones, using limited labeled data and demonstrations.
Owing to the modular design of the skill graph, newly added skills do not interfere with existing ones—unlike monolithic foundation models, which often degrade when updated.

\subsection{Limitations \& Future Works}
As shown in \cref{tab:object_successes,tab:wbcd-success}, our system encountered several failures during both development and the competition. A key challenge during development was the limited reachability of the G1 arms—some objects were placed beyond the robot’s reachable workspace, leading to failed grasps. This is partly due to the fact that our skill graph is constructed in Cartesian space, where waypoints are annotated with semantic labels (e.g., required vs. optional), and later mapped back to the configuration space. However, this mapping is not guaranteed to be complete or optimal; in some cases, feasible configuration-space motions may not exist for a given Cartesian plan. While the approach works well in practice, these limitations could be mitigated by enabling whole-body control to expand the reachable workspace. During the competition, the main source of failure shifted to perception: the limited field of view (FOV) of the wrist cameras often led to incomplete single-shot observations, resulting in perception errors and reduced grasp success rates. Incorporating wider-FOV cameras or multi-view fusion techniques may alleviate this issue. Future work will investigate how to assess the feasibility of Cartesian motion plans under configuration-space constraints and how to enhance perception robustness for more reliable execution.





\bibliographystyle{plainnat}
\bibliography{references}

\end{document}